\documentclass[letterpaper, 10 pt, conference]{ieeeconf}  % Comment this line out if you need a4paper

\IEEEoverridecommandlockouts                              % This command is only needed if 
\usepackage[latin1,utf8]{inputenc}
\usepackage{graphics} % for pdf, bitmapped graphics files
\usepackage{epsfig} % for postscript graphics files
\usepackage{mathptmx} % assumes new font selection scheme installed
\usepackage{times} % assumes new font selection scheme installed
\usepackage{amsmath} % assumes amsmath package installed
\usepackage{amssymb}  % assumes amsmath package installed

\usepackage{multirow}
\usepackage{multicol}
\usepackage{graphicx} % DO NOT CHANGE THIS
\usepackage{xcolor} %TXT Color 
\usepackage{array}
\usepackage{subfigure}
\usepackage{leftidx}
\usepackage{float}
\usepackage{amsmath}
\usepackage{amssymb}
\usepackage[ruled,vlined]{algorithm2e}
\usepackage{lineno}
\usepackage{hyperref}
\usepackage{threeparttable}

\title{\LARGE \bf
Faraway-Frustum: Dealing with Lidar Sparsity for 3D Object Detection using Fusion
}

\author{Haolin Zhang*, Dongfang Yang*, Ekim Yurtsever*, Keith A. Redmill and Ümit Özgüner
\thanks{*Equal contribution}% <-this % stops a space
\thanks{Authors are with the Department of Electrical and Computer Engineering, The Ohio State Universtiy, OH 43210, USA.
        Contact: {\tt\small [zhang.10749,yang.3455,yurtsever.2,
        redmill.1,ozguner.1] @osu.edu}}%
}

\begin{document}

\maketitle
\thispagestyle{empty}
\pagestyle{empty}

%%%%%%%%%%%%%%%%%%%%%%%%%%%%%%%%%%%%%%%%%%%%%%%%%%%%%%%%%%%%%%%%%%%%%%%%%%%%%%%%
\begin{abstract}

Learned pointcloud representations do not generalize well with an increase in distance to the sensor. For example, at a range greater than 60 meters, the sparsity of lidar pointclouds reaches to a point where even humans cannot discern object shapes from each other. However, this distance should not be considered very far for fast-moving vehicles: A vehicle can traverse 60 meters under two seconds while moving at 70 mph. For safe and robust driving automation, acute 3D object detection at these ranges is indispensable. Against this backdrop, we introduce faraway-frustum: a novel fusion strategy for detecting faraway objects. The main strategy is to depend solely on the 2D vision for recognizing object class, as object shape does not change drastically with an increase in depth, and use pointcloud data for object localization in the 3D space for faraway objects. For closer objects, we use learned pointcloud representations instead, following state-of-the-art. This strategy alleviates the main shortcoming of object detection with learned pointcloud representations. Experiments on the KITTI dataset demonstrate that our method outperforms state-of-the-art by a considerable margin for faraway object detection in bird's-eye-view and 3D. Our code is open-source and publicly available: \url{https://github.com/dongfang-steven-yang/faraway-frustum}.

\end{abstract}

%%%%%%%%%%%%%%%%%%%%%%%%%%%%%%%%%%%%%%%%%%%%%%%%%%%%%%%%%%%%%%%%%%%%%%%%%%%%%%%%
\section{INTRODUCTION}
%%%%%%%%%%%%%%figure1

3D/Bird's eye view (BEV) object detection is a critical task for many robotics applications. Existing lidar-based methods show good performance for close to medium range objects. However, a closer look at the state-of-the-art exposes an inherent problem: learned pointcloud representations do not generalize well with an increase in sparsity. This is not a surprising phenomenon. At a range greater than sixty meters, lidar pointcloud sparsity reaches a point where even humans cannot discern object shapes from each other. For example, in KITTI 3D/BEV object detection benchmark \cite{kt}, the state-of-the-art 3D object detection performance is remarkable. But when these high performing models face objects that are located at 60 meters and beyond, mean average precision drops to almost \textit{zero}. We believe this is an important issue for automated driving. For instance, detecting faraway objects can offer more time for the automated vehicle to make better decisions.

% Early works for faraway objects mainly focused on 2D instead of 3D/BEV~\cite{pdfar1,pdfar2}. Nowadays, 2D detection of faraway objects has achieved very good and robust results with the development of 2D vision approaches~\cite{fasterrcnn,maskrcnn}. 

\begin{figure}[!ht]
\begin{center}{\includegraphics[width=0.9\linewidth]{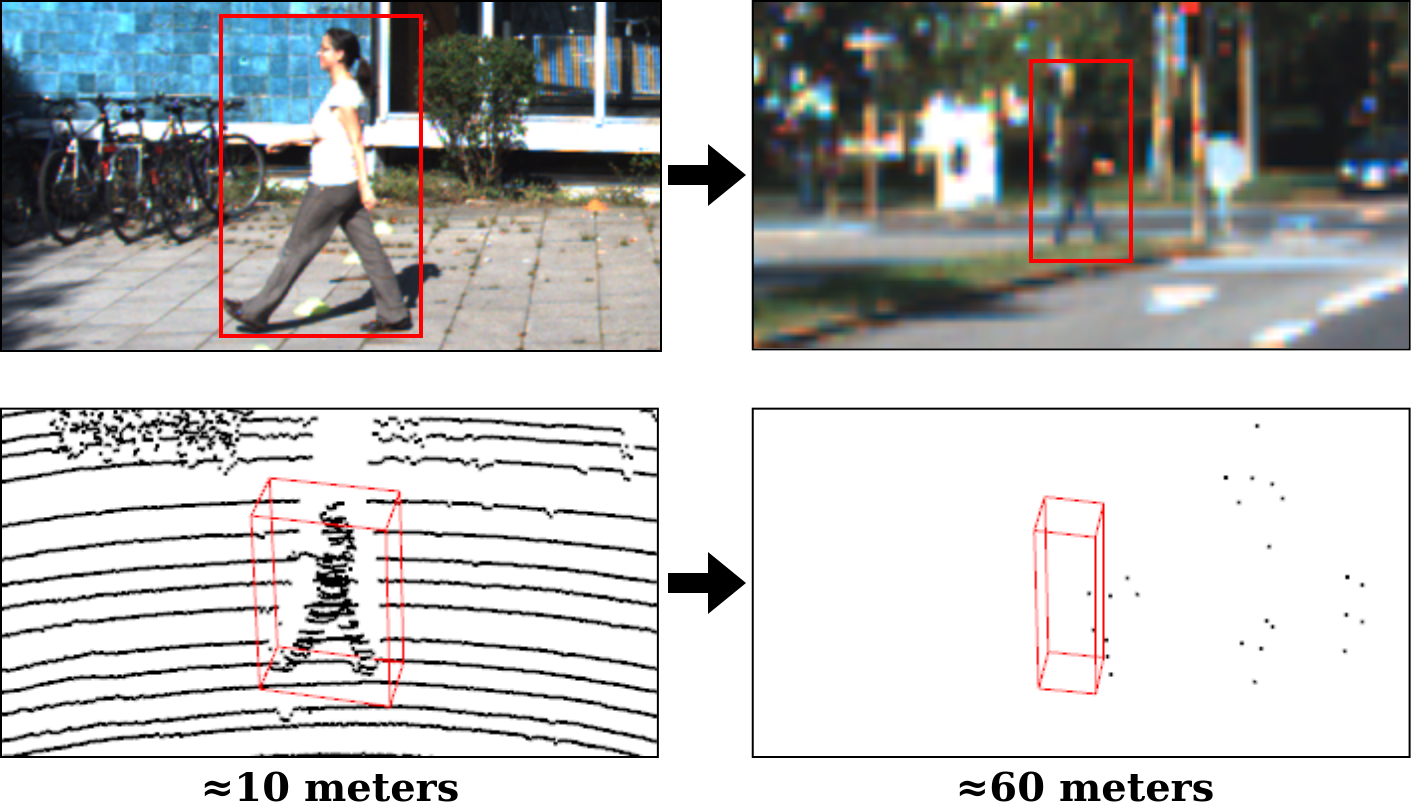}}
\end{center}
\vspace{-0.2cm}
   \caption{Learned pointcloud representations do not generalize well with an increase in sparsity. This problem does not translate to the 2D RGB image domain in the same fashion, as object shape does not change drastically with an increase in depth. However, sparse points in the target object's vicinity can still be used to estimate depth. Our method utilizes these sparse points to estimate depth while using 2D RGB information to recognize shape and object-class.}
\label{fig:v1}
\vspace{-0.5cm}
\end{figure}

3D/BEV object detection for faraway objects is challenging, and state-of-the-art (SOTA) lidar-based detectors~\cite{pvrcnn, second, pointpillars, tanet, 3dssd} do not perform well for this task. We believe this is caused by sparsity and near-random scattering of the few points obtained from faraway objects. Learned representations from close to medium range objects do not generalize to faraway cases, and since SOTA approaches are primarily deep neural networks, they cannot learn the representations of faraway cases. 

RGB-pointcloud fusion is a common strategy  \cite{fp, amodal, MV3D, AVOD, slide, fc} to increase 3D object detection performance. For example, some works~\cite{fp}, \cite{fc} focus on using 2D detection results to generate frustum-based search spaces in pointclouds. As shown in Fig. \ref{fig:v1}, a faraway object in the RGB image domain usually contains around 400 pixels, which can be easier to recognize with a mature 2D detector. As such, a fusion-based approach can be a good candidate for faraway object detection. However, even though the aforementioned studies use RGB imagery to boost detection performance, they still depend on solely learned pointcloud representations to localize objects in 3D.  

%Even though a recent work \cite{LRPD} introduced a features-fusion strategy in the frustum-based approach to improve the performance of detecting pedestrians over 30 meters, to the best of our knowledge, there is no work concerning objects at extremely faraway distances (e.g. over 60 meters).

\begin{figure*}[htb]
\begin{center}{\includegraphics[width=1\linewidth]{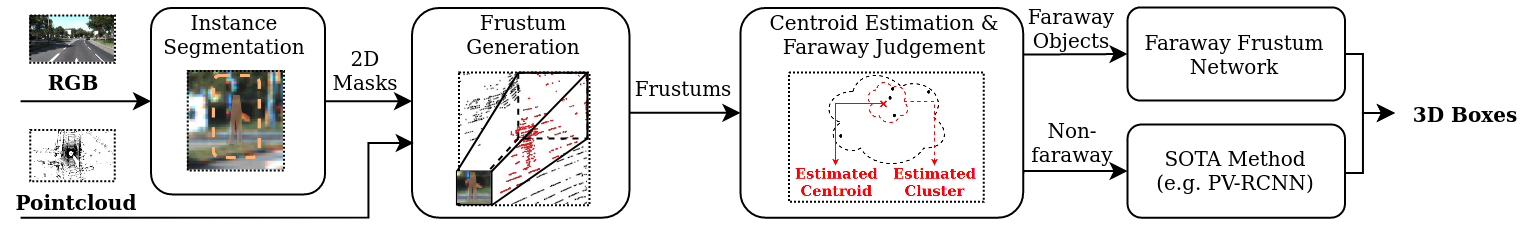}}
\end{center}
   \caption{Overview of the 3D/BEV object detection system based on our proposed method ({\textit{Faraway-Frustum}}). It contains three main stages: frustum generation, centroid estimation, and box regression. First, the 2D object information (classification and 2D semantic mask) is extracted from the image by conducting instance segmentation, and then the 3D frustum is shaped by extruding the 2D semantic mask to the 3D coordinate system. Second, lidar pointcloud (red points) in the frustum are collected and clustered, and then the 3D object centroid is estimated. Finally, depending on the faraway judgment, the 3D bounding box is predicted by our Faraway Frustum Network or a state-of-the-art method.}
\label{fig:pipeline}
\end{figure*}

In this work, we propose an alternative 3D/BEV detector, \textit{Faraway-Frustum}, to address the problem of faraway object detection. We follow the idea of frustum generation but use clustering instead of a neural network to estimate an initial object location in the cropped pointcloud for faraway objects. We still train a neural network to regress bounding box shape and refine depth. The overview of the proposed method is shown in Fig.~\ref{fig:pipeline}. We first use 2D instance segmentation masks (or 2D bounding boxes) for each object in the RGB image space to generate frustums in the pointcloud space and find the corresponding lidar points for each object. Then, a pointcloud clustering technique is applied to estimate the 3D centroid of the object. By comparing the centroid with a faraway threshold, an object is determined to be treated as a faraway object or not. If so, a 3D bounding box is regressed by our Faraway Frustum Network (FF-Net) to the object based on the estimated centroid and frustum pointcloud . If it is not a faraway object, instead of clustering the raw pointcloud, learned representations are directly used for 3D box fitting, following SOTA. 

To evaluate the proposed method, we conducted bench-marking experiments using the KITTI dataset~\cite{kt}. In KITTI, the average number of lidar points for each faraway object (e.g. pedestrians over 60 meters and cars over 75 meters) is ten or less, which supports our motivation that an alternative approach is necessary for faraway objects instead of directly using pointcloud-driven neural network approaches. The experimental results demonstrated that our method outperforms the SOTA methods on faraway object detection, which indicates that our method effectively fuses the RGB data with very sparse pointcloud. As shown in Fig. \ref{fig:example}, our proposed method successfully detects faraway objects, where SOTA methods (fusion or pointcloud only) fail. Our main contributions can be summarized as follows:

\vspace{0.2cm}

%%%%%%%%%%%%%%%%%%%%%%%%%%%%%%%%%contributions

$\bullet$ Introduction of a novel fusion strategy: depending solely on 2D vision for object-class recognition and using frustum-cropped pointcloud data with clustering for 3D object localization. 

$\bullet$ Showing that using clustering with cropped, very sparse raw pointcloud data is a better strategy than using learned representations for faraway 3D object detection. As shown in Fig. \ref{fig:v1}, within very sparse pointclouds, the shape of objects changes drastically and randomly. As such, using representations learned mostly from closer objects is not useful.  

$\bullet$ Demonstrating state-of-the-art 3D object detectors' failure with objects at a distance over sixty meters in the KITTI dataset. The proposed faraway-frustum approach outperforms SOTA with a significant margin.

%%%%%%%%%%%%%%%%%%%%%%%%%%%%%%%%%%%%%%%%%%%%%%%%%%%%

%%%%%%%%% Related Work
\section{RELATED WORK}

In this section, we briefly review state-of-the-art 3D/BEV object detection methods. We divide them into two main categories: pointcloud only methods and RGB-pointcloud fusion methods. In the second category, we mainly discuss feature-based fusion and frustum-based fusion. We also discuss their performance of detecting faraway objects.

{\bf Pointcloud only methods.} One way of processing pointcloud is based on voxels~\cite{second,pointpillars,pvrcnn,hvnet}. Such methods first convert the pointcloud into voxel grids and then learn the representation of each voxel. 3D/BEV detection is achieved with the learned voxel representation. Alternatively, raw pointcloud data can be used directly utilizing PointNet-based architectures \cite{pointnet} and make 3D/BEV object detections~\cite{pointrcnn,std,3dssd}. These methods are robust for most objects. However, pointcloud only methods all have the difficulty of detecting faraway objects, because lidar points of faraway objects are too sparse to be voxelized and learned, leading to no detection result in most cases.

{\bf Feature-based Fusion.} Feature-based fusion methods try to make the pointcloud data and the RGB data complement each other. One way is to fuse the information from pointcloud into RGB image. For example, \cite{amodal} fuses the features from Region of Interest (RoI) in both 2D image and 2D depth map, and then conducts 3D box regression. MV3D~\cite{MV3D} projects the lidar pointcloud to two-view image representations (Bird view and Front view). The features and information extracted from these two image representations and the RGB image are then fed into a region-based fusion network for 3D object detection. AVOD~\cite{AVOD} firstly generates a BEV map from a voxel grid representation of the lidar pointcloud. The features extracted from both the BEV map and the RGB image are fused for 3D object detection through a first-stage region proposal network and a second-stage detector network. The main problem of these methods is the loss of 3D geometric information of lidar pointcloud because of using the pointcloud's 2D representations, leading to some errors in locating small objects such as pedestrians. Feature-based fusion can also be achieved by fusing the information from image space into pointcloud. For example, \cite{slide} extracts the geometric features in 3D and color features in 2D from RGB-D images and then fuses them for 3D object detection. MVX-Net~\cite{mvnet} fuses the RGB image and pointcloud point-wise or voxel-wise. The features extracted from the RGB image by a pre-trained 2D CNN are fused with the pointcloud in a voxel-based network to do 3D object detection.
PointPainting~\cite{pointpainting} assigns the semantic feature to each lidar point by fusing the 2D detection result from the RGB image, thus achieving better results in pointcloud-based neural network detector. Since these approaches heavily rely on the pointcloud feature, they still can not generate good results for faraway objects with sparse lidar points. 

{\bf Frustum-based Fusion.} Frustum-based fusion methods use the detection results from 2D image to generate frustums for pointcloud, hence reducing the search space in 3D. An early and classic method is Frustum PointNets~\cite{fp}. This method first generates a frustum for each object detected in 2D, then applies a PointNet-based approach to do instance segmentation and 3D box estimation in each frustum. 
Some work \cite{frustum2,frustum3} have improved the process of frustum generation by filtering out some background noise, and there is work focusing on changing the content of frustums. For example, Frustum ConvNet~\cite{fc} generates a sequence of sub-frustums via sliding in the original 3D frustum. Frustum Voxnet~\cite{frgbd} voxelizes parts of the frustum instead of using the whole frustum space, which offers more accurate representations around the area of interest. 
Some other researchers aimed to provide more fusion information. For example, one work~\cite{frustum6} combines the pointcloud features in the frustum with the image features in the 2D bounding box as early-fusion and then apply a PointNet-based detector. Another work~\cite{multif} fuses their own BEV detection results with 3D/BEV results from Frustum PointNets as late-fusion. 
These perform well for most objects. But unfortunately, for faraway objects with sparse lidar points, pure neural network based approaches cannot generalize well. 

One recent work~\cite{LRPD} achieved good 3D/BEV pedestrian detection results around 30 meters. In our work, we extend the range significantly, and detect pedestrians at 60 meters and beyond, where most SOTA approaches completely fail.

%-------------------------------------------------------------------------
%-------------------------------------------------------------------------

%%%%%%%%% Our method
\section{PROPOSED METHOD}

\begin{figure*}[htb]
\begin{center}{\includegraphics[width=0.8\linewidth]{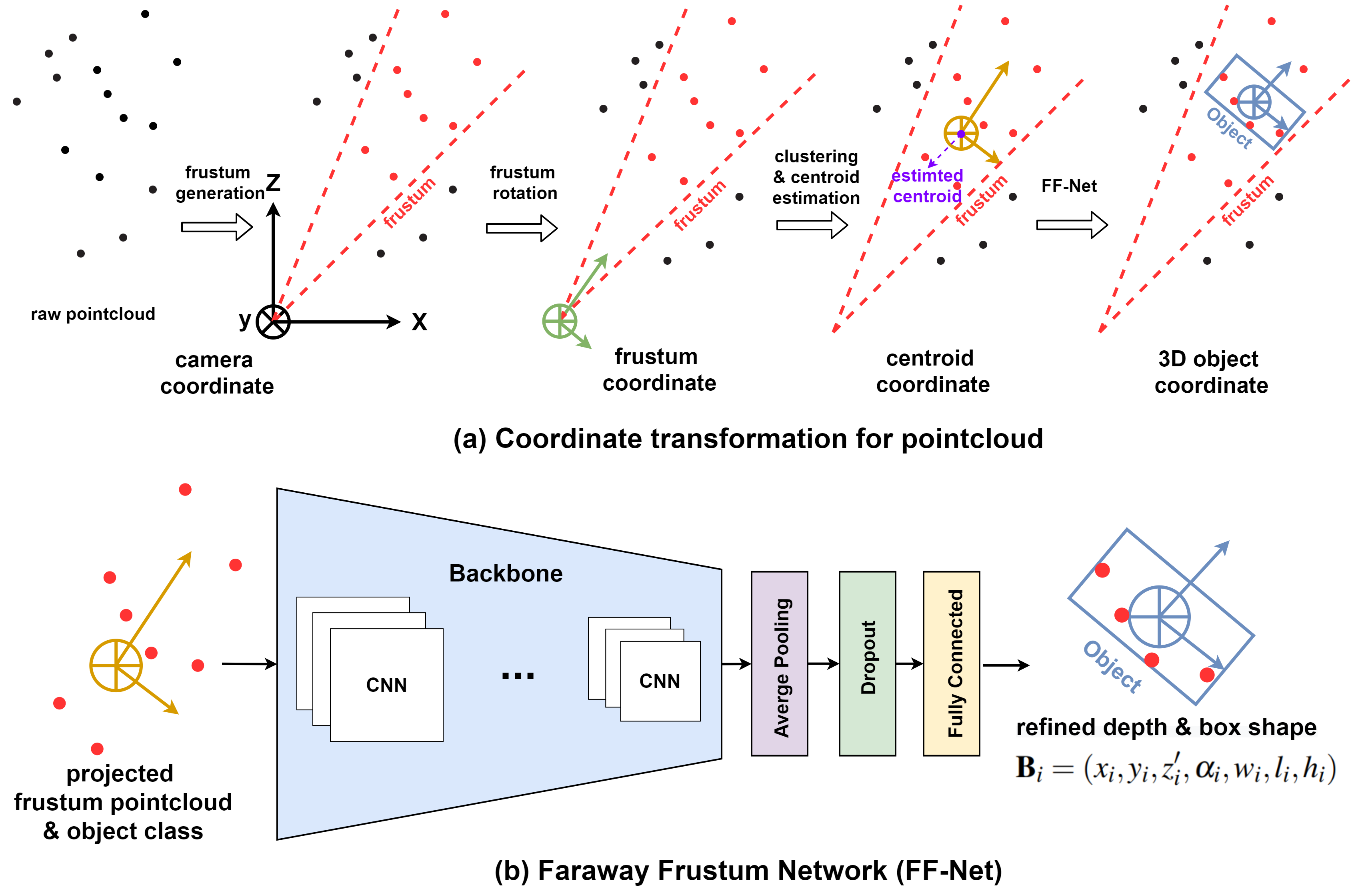}}
\end{center}
\caption{An illustration of coordinate transformation for pointcloud and Faraway Frustum Network (FF-Net). (a) Coordinate transoformation for pintcloud shows the process of projecting pointcloud into different coordinate systems in our method. After carrying out frustum generation, frustum rotation,  clustering and centroid estimation, frustum pointcloud is projected into the centroid coordinate system. Our goal is to further localize the 3D object by the 2D projection of the frustum pointcloud and the FF-Net. (b) The FF-Net is essential to refine the object center, regress box size, and resolve certain issues that may occur while creating the frustum. For example, due to errors in detection or segmentation, the cluster centroid may not be aligned well with the object. The FF-Net is trained to deal with such issues and refine object localization. }
% given two objects where their frustrums are overlapping, and the closer object is either included in the frustrum of the further object, or due to errors in detection or segmentation, the closer object is visible in the scene. The FF-Net is trained to deal with such issues.}
\label{fig:ffnet}
\end{figure*}

An overview of our proposed method is shown in Fig.~\ref{fig:pipeline} and Algorithm~\ref{algo}. Our method takes both the RGB image and lidar pointcloud as input and outputs 3D/BEV bounding boxes $\mathbf{B}_i$ with class id $c_i$. There are three main stages in our method: frustum generation, centroid estimation, and depth-refinement with box regression. Each stage will be illustrated in detail in the following subsections.

\subsection{Frustum Generation}

{\bf 2D instance segmentation.} 2D instance segmentation serves as the base of frustum generation. It takes an image as input and outputs the 2D object detection results containing 2D bounding boxes and semantic masks.

In this work, we use 2D instance segmentation framework Mask R-CNN~\cite{maskrcnn} to obtain 2D object information $\{R_i\}$ from image $\mathbf{I}$:
\begin{equation}
    \{R_i\}=f_\text{Mask R-CNN}(\mathbf{I})
\end{equation}
where $f_\text{Mask R-CNN}$ represents the Mask R-CNN framework. $R_i=(c_i, \mathbf{b}_i,\mathbf{M}_i,s_i)$ is the instance segmentation result, which is a 4-tuple consisting of class label $c_i$, 2D bounding box $\mathbf{b}_i$, 2D semantic mask $\mathbf{M}_i$, and confidence score $s_i$ for object $i$.

\begin{figure}[t]
\centering
\includegraphics[width=0.8\columnwidth]{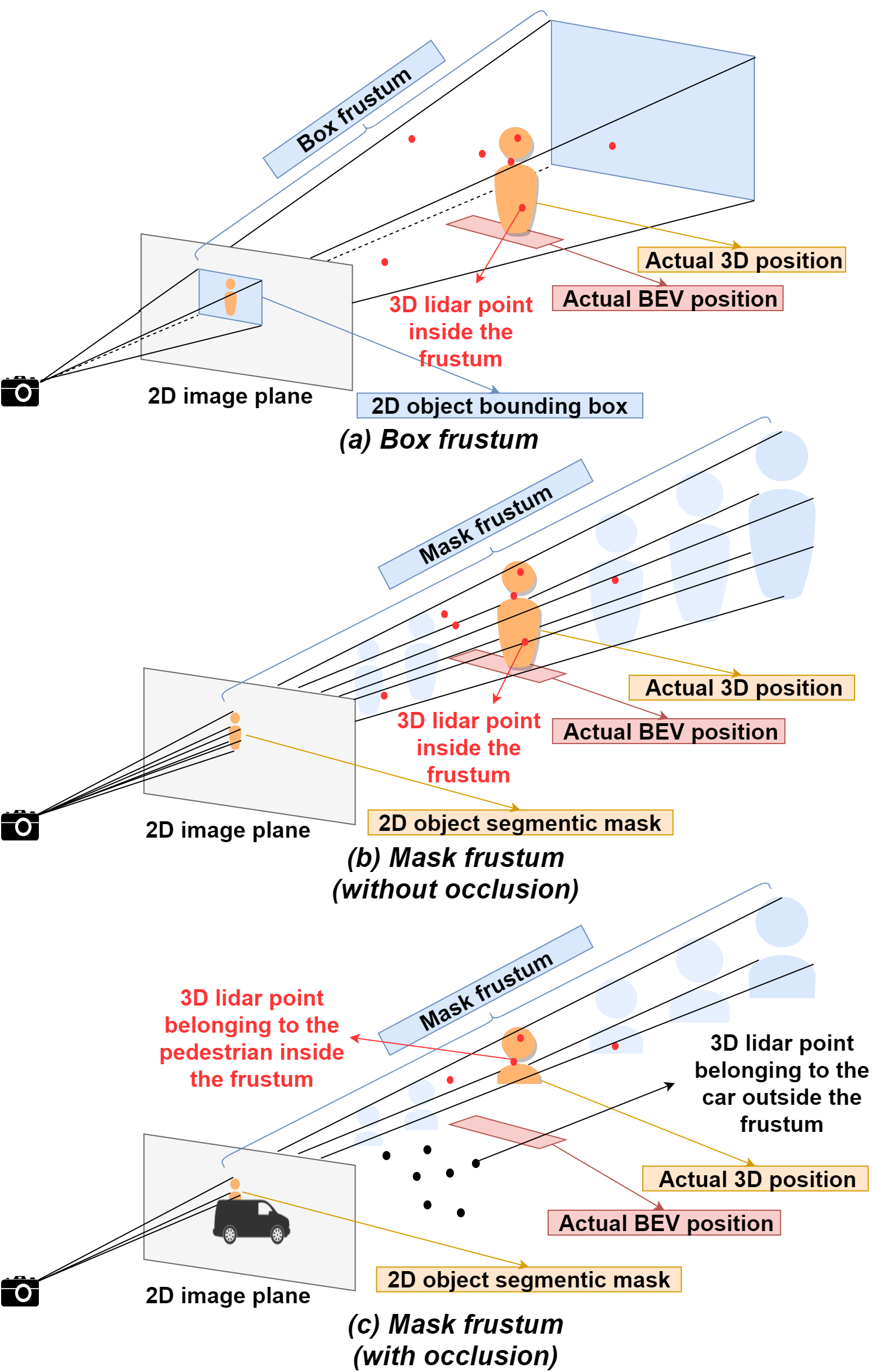} % Reduce the figure size so that it is slightly narrower than the column. Don't use precise values for figure width. This setup will avoid overfill boxes.
\caption{An illustration of frustum generation. The main difference between box frustum and mask frustum is that box frustum uses the 2D bounding box as the projection source, while mask frustum uses the 2D semantic mask. Mask frustum gives the more compact search space alongside the outline of the object, and thus excludes some noise points caused by potential occlusions.}
\label{fig:maskf}
\end{figure}

{\bf Frustum generation.} We use the 2D results $\{R_i\}$ to generate frustums and to further identify the lidar points that correspond to each object $i$. With the known transformation $\mathbf{T}$ between the camera and the lidar, we use the semantic mask $\mathbf{M}_i$ for 2D-to-3D projection, as shown in Fig.~\ref{fig:maskf}(b). Then the corresponding pointcloud $\mathbf{P}_{i}^{\prime}$ can be identified from the raw lidar pointcloud $\mathbf{P}$ based on the frustum:
\begin{equation}
    \mathbf{P}_{i}^{\prime}=f_\text{Mask-frustum}(\mathbf{M}_i, \mathbf{T}, \mathbf{P})
\end{equation}
The mask-frustum based projection $f_\text{Mask-frustum}$ is our main approach. As shown in Fig.~\ref{fig:maskf}(c), we believe that using the semantic mask can exclude some noise points that do not belong to the target object, e.g., the points from occluded objects or the background. As an alternative, we also tested box-frustum based projection $\mathbf{P}_{i}^{\prime}=f_\text{Box-frustum}(\mathbf{b}_i, \mathbf{T}, \mathbf{P})$, which is used as a comparison with our main approach.

\begin{algorithm}[t]
\SetAlgoLined
\KwIn{\\ Lidar pointcloud $\mathbf{P} \in \mathbb{R}^{N,3}$.\\
RGB image $\mathbf{I} \in \mathbb{R}^{H, W,3}$. \\
Calibration matrix $\mathbf{T}\in \mathbb{R}^{4,4}$. \\
Faraway object threshold $z_\text{th} \in \mathbb{R} $
}

\KwOut{\\ 3D object bounding box $\mathbf{B}_i\in \mathbb{R}^{7}$.\\
Class id $c_i$.}
 {\bf Main algorithm:} \\
 \{$c_i, \mathbf{b}_i,\mathbf{M}_i,s_i \}=f_\text{Mask R-CNN}(\mathbf{I})$ ($i=1,2...n$)
%  ${c_i, \mathbf{b}_i,\mathbf{M}_i,s_i}=f_\text{Mask R-CNN}(\mathbf{I})$ ($i=1,2...n$)\\
\;
\ForEach{$i(1,2...n)$}{
$\mathbf{P}_{i}^{\prime}=f_\text{Mask-frustum}(\mathbf{M}_i, \mathbf{T}, \mathbf{P})$\;
$(x_i, y_i, z_i)=f_\text{clustering}(\mathbf{P}_{i}^{\prime})$\;
\eIf{$z_i \geq z_\text{th}$}{
      $ \mathbf{P}_{i}^{\prime \prime} = f_\text{projection}(\mathbf{P}_i^{\prime}, x_i, y_i, z_i)$\;
      $(z_i^{\prime},w_i, l_i, h_i,\alpha_i) = f_\text{FF-Net}(\mathbf{P}_{i}^{\prime \prime},c_i)$\;
    $\mathbf{B}_i=(x_i, y_i, z_i^{\prime}, \alpha_i, w_i, l_i, h_i)$\;
    }{
     $\mathbf{B}_i, c_i = f_\text{SOTA} (\mathbf{P},\mathbf{I})$\;
    }
}
 \caption{Faraway-Frustum({$\mathbf{P}$, $\mathbf{I}$, $\mathbf{T}$, $z_\text{th}$})}
 \label{algo}
\end{algorithm}

\subsection{Centroid Estimation}

{\bf Centroid estimation.} With $\mathbf{P}_{i}^{\prime}$ obtained from the frustum, we then estimate the 3D centroid $(x_i, y_i, z_i)$ for object $i$:
\begin{equation}
    (x_i, y_i, z_i)=f_\text{clustering}(\mathbf{P}_{i}^{\prime})
\end{equation}
The 3D object centroid plays two key roles in our method. One is to use the depth $z_i$ to determine whether object $i$ should be treated as a faraway object. The other is to further generate the 3D/BEV detection results for faraway objects. 

Based on our observation in KITTI dataset, no matter whether the pointcloud in the frustum is dense or sparse, there are always some points on the object's surface. Thus, we adopt a fast clustering technique using histograms to estimate the 3D object centroid.

First, for all points in the pointcloud $\mathbf{P}_{i}^{\prime}$, the histogram of all the coordinate values in each axis is generated (here we have 3 axes x, y, and z). For the histogram of each axis, we define the edges of every bin in the histogram as $(e^{l}_{j},e^{r}_{j})$, $\forall j \in \{0,1,\cdots,N \}$, and the count of values belonging to each bin as $n_j$, $\forall j \in \{0,1,\cdots,N \}$, where $N$ is the number of bins. 
Then, we identify the bin with the largest count value. This indicates that most of the points are concentrated within this bin. The corresponding index will be obtained by $j^* =\arg\max_{j}{(n_j)}$.
Finally, the centroid value of an axis, for example, the centroid of x-axis, $x_i$, can be obtained by $ x_i = \frac{1}{2}(e^{l}_{j^*}+e^{r}_{j^*}).$
The centroid values $y_i$ and $z_i$ for the other two axes is estimated in the same way.

\begin{figure}[t]
\centering
\includegraphics[width=0.9\columnwidth]{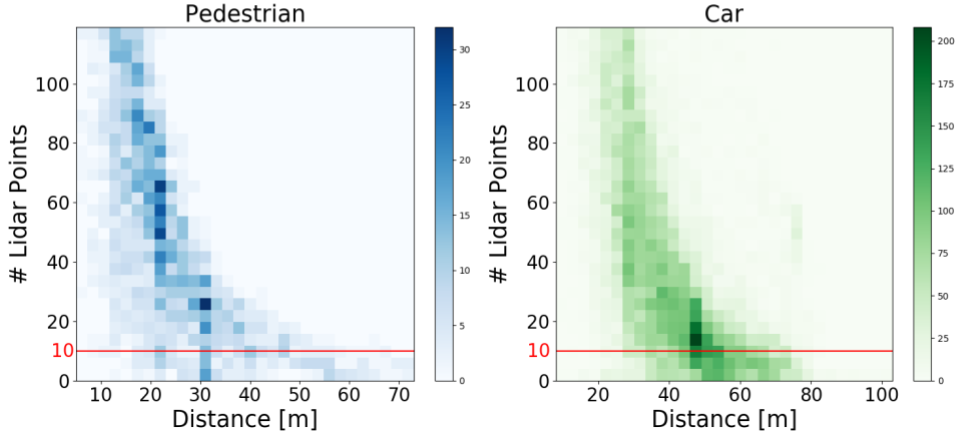} % Reduce the figure size so that it is slightly narrower than the column. Don't use precise values for figure width. This setup will avoid overfill boxes.
\caption{The number of points belonging to an object (pedestrians and cars) versus distance from the sensor in the KITTI dataset. As the distance (x-axis) increases, the number of lidar points in an object (y-axis) decreases drastically. This indicates that the pointcloud of each faraway object is very sparse. We use this distribution to decide the faraway decision threshold.}
\label{fig:map}
\end{figure}

{\bf Faraway threshold.} Using the estimated centroid $(x_i, y_i, z_i)$, we conduct faraway judgement for each object $i$. We set different faraway thresholds $z_\text{th}$ for different object classes based on the statistics of the number of ground truth lidar points in each object. As shown in Fig.~\ref{fig:map}, we first draw a line for the objects
that have 10 lidar points, then we approximately select the $z_\text{th}$ such that most objects of distance larger than $z_\text{th}$ have less than 10 points. To determine whether object $i$ should be treated as a faraway object, we compare the estimated distance $z_i$ with $z_\text{th}$ of the corresponding object class $c_i$. If $z_i>z_\text{th}$, then it is a faraway object, otherwise not.
% We set different faraway thresholds $z_\text{th}$ for different object classes (pedestrian: 60 meters, car: 75 meters).
\subsection{Box Regression}

To obtain the 3D/BEV bounding box $\mathbf{B}_i$ for object $i$, based on the estimated object centroid $(x_i, y_i, z_i)$, we need to estimate the box shape: length $l_i$, width $w_i$, height $h_i$, and the orientation $\alpha_i$. If object $i$ is a faraway object, directly using learned representations from state-of-the-art models is not a good choice because they do not generalize well from dense pointcloud to very sparse pointcloud. Furthermore, the estimated object centroid (especially the depth) may still be quiet far from the box center. As such, we propose to use a light model named Faraway Frustum Network (FF-Net) only for faraway objects to refine the depth $z_i^{\prime}$ and regress the shape $(w_i, l_i, h_i, \alpha_i)$ of the 3D bounding box with the input of the object class $c_i$ and a 2D projection $ \mathbf{P}_{i}^{\prime\prime}$ of the frustum pointcloud, as shown in Fig.~\ref{fig:ffnet}. 

{\bf Pointcloud Projection.} 2D projection $ \mathbf{P}_{i}^{\prime\prime}$ of the frustum pointcloud  is generated from cooridinate transformation for pintcloud, as shown in Fig.~\ref{fig:ffnet} (a). After conducting frustum generation for raw pointcloud $\mathbf{P}$, the frustum pointcloud $ \mathbf{P}_{i}^{\prime}$ is obtained. First, the camera coordinate system is rotated to the center view of the frustum to build frustum coordinate system. Second, after histogram-based clustering and centroid estimation, the frustum coordinate system is transformed to the centroid coordinate system with the estimated centroid at origin. At last, lidar points inside of the frustum are all projected into the centroid coorinate system in bird's eye view. The projected frustum pointcloud in centroid coordinate system is taken as the 2D projection $ \mathbf{P}_{i}^{\prime\prime}$ of the frustum pointcloud.

{\bf Box Regression.} FF-Net takes the object class $c_i$ and a 2D projection $ \mathbf{P}_{i}^{\prime\prime} = f_\text{projection}(\mathbf{P}_{i}^{\prime}, x_i, y_i, z_i)$  of the frustum pointcloud $\mathbf{P}_{i}^{\prime}$ whose origin is the estimated centroid $(x_i, y_i, z_i)$ as input, and combines a MobileNet-based~\cite{mobile} backbone network with a multi-output regression head as shown in Fig. \ref{fig:ffnet} (b).

% \begin{figure}[htb]
% \begin{center}{\includegraphics[width=1\linewidth]{img/project4.png}}
% \end{center}
%   \caption{projection.}
% \label{fig:projection}
% \end{figure}

% \begin{figure}[htp]
% \centering
% \includegraphics[width=1\columnwidth]{img/ffnet9.png} % Reduce the figure size so that it is slightly narrower than the column. Don't use precise values for figure width. This setup will avoid overfill boxes.
% \caption{An illustration of Faraway Frustum Network (FF-Net). The FF-Net is essential to refine the object center, regress box size, and resolve certain issues that may occur while creating the frustum. For example, imagine there are two objects where their frustrums are overlapping. And the closer object is either included in the frustrum of the further object, or due to errors in detection or segmentation, the closer object is visible in the scene. The FF-Net is trained to deal with such issues.}

% % We still need a final FF network to refine the objcet center, regress box size and resolve certain issues that can occur while creating the frustum. For example, imagine there are two objects where their frustrum is overlapping. And the closer object is either included in the frustrum of the farther object or due to errors in detection or segmentation the closer object is visible in the scene.

% \label{fig:ffnet}
% \end{figure}

The estimated centroid $(x_i, y_i, z_i)$ is considered as the origin of the input projection, and the goal of the FF-net is to shift this origin to the real center of the 3D bounding box (i.e. to transform centroid coordinate to 3D object coordinate as shwon in Fig.~\ref{fig:ffnet} (a)). Furthermore, another goal is to regress the box shape, achieved by minimizing the loss of the regressed length, width and height of the box. We use mean absolute error (MAE) to compute the loss $L_x$,$L_y$,$L_z$,$L_w$,$L_l$,$L_h$,$L_\alpha$ of box centroid $(x_i^{\prime}, y_i^{\prime}, z_i^{\prime}$) and shape ($w_i, l_i, h_i,\alpha_i)$ respectively. By summing these losses, FF-Net is trained and optimized with multi-task losses $L_{FF-Net}$.
% \begin{equation}
%     \mathcal{L}_{FF-Net} = \mathcal{L}_x + \mathcal{L}_y + \mathcal{L}_z + \mathcal{L}_w + \mathcal{L}_l + \mathcal{L}_h +\mathcal{L}_\alpha
% \end{equation}

Finally, for a faraway object, we take the shifted depth $z_i^{\prime}$ and the regressed 3D bounding box shape $(w_i, l_i, h_i, \alpha_i)$ from the output of FF-Net and combine them with ($x_i,y_i$) from the estimated centroid. We assign a 3D bounding box to the faraway object $i$ as:
%Finally, for a faraway object, we take the shifted depth $z_i^{\prime}$ and the regressed 3D bounding box shape $(w_i, l_i, h_i, \alpha_i)$ from the ouput of FF-Net, as $(z_i^{\prime},w_i, l_i, h_i,\alpha_i) = f_\text{FF-Net}(\mathbf{P}_{i}^{\prime}^{\prime},c_i)$. Combining them with ($x_i,y_i$) from the estimated centroid, we assign a 3D bounding box to the faraway object $i$ (the object class $c_i$ is obtained from 2D instance segmentation) with:
\begin{equation}
    \mathbf{B}_i=(x_i, y_i, z_i^{\prime}, \alpha_i, w_i, l_i, h_i).
\end{equation}

It should be noted that the class id $c_i$ is directly obtained with Mask R-CNN. If object $i$ is not a faraway object, we switch to using learned representations, following SOTA (e.g. Frustum-PointNets \cite{fp}, PV-RCNN \cite{pvrcnn}). In this case, 3D bounding box and class id for a non-faraway object are obtained by $\mathbf{B}_i, c_i = f_\text{SOTA} (\mathbf{P},\mathbf{I})$.

\section{EXPERIMENTS}

% To evaluate our method's performance, we compare our method with state-of-the-art models on the KITTI dataset and benchmark \cite{kt}. 
We utilized the KITTI dataset~\cite{kt} to conduct our experiments. We specifically extracted faraway objects in KITTI and investigated them separately. Details of dataset preparation, evaluation metrics, and implementation are described below.

{\bf Dataset preparation.} First we analyzed the statistics of the original KITTI dataset by evaluating the distribution of the objects at different distances and having different numbers of lidar points, as shown in Fig. \ref{fig:map}. It is obvious that as the distance increases, the number of lidar points in an object decreases. That is to say, the pointcloud is very sparse for faraway objects. We selected the faraway threshold $z_\text{th}$ for cars as 75 meters and pedestrians as 60 meters. We also split the KITTI dataset into the train set (3724 frames) and the validation set (3757 frames). %According to our faraway thresholds setting, we have 29 pedestrians at over 60 meters and 75 cars at over 75 meters in KITTI val set. 

\begin{figure*}[htb]
\begin{center}{\includegraphics[width=1\linewidth]{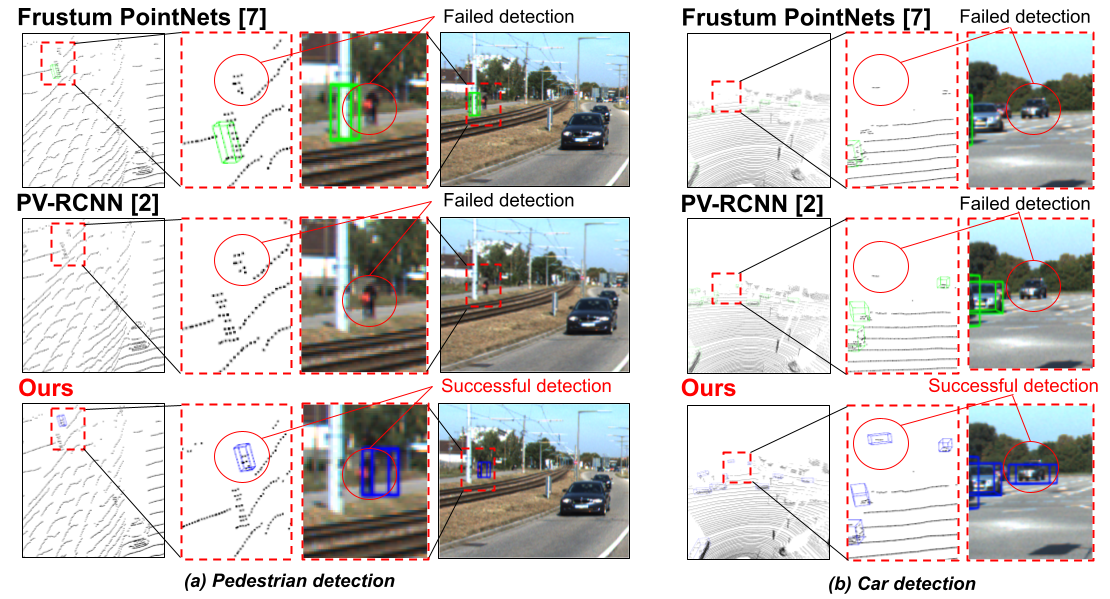}}
\end{center}
   \caption{Example 3D detection results of faraway objects from the KITTI test set. (a) Pedestrian detection. \textit{Top row}: Frustum PointNets \cite{fp}, which is based on fusing multiple modalities (RGB and pointcloud).  \textit{Middle row}: PV-RCNN \cite{pvrcnn}, which is based on only pointcloud.  \textit{Bottom row}: Our proposed method. (b) Car detection. Same arrangement as in (a). In these examples, for both faraway pedestrian and faraway car, the proposed method (Ours) successfully detects the targets. However, state-of-the-art methods (Frustum PointNets~\cite{fp} and PV-RCNN~\cite{pvrcnn}) all fail.}
\label{fig:example}
\end{figure*}

{\bf Evaluation metric.} The major evaluation metric is the mean average precision (mAP) with a given IoU threshold, as suggested by the KITTI dataset. We use both the official benchmark and a specific benchmark for faraway objects. In the benchmark for faraway objects, we only evaluate faraway objects and we use a specific IoU threshold (0.1) for the mAP. We use low IoU threshold because detecting the object with a small overlap with the ground truth is still helpful at long distances, compared to no detection at all. The mAP results are computed with 11 recall positions which is the same as in previous works.

We also evaluate the faraway objects using average IoU (\textit{aIoU}), which is defined as $\textit{aIoU}=\frac{\sum_{i=1}^n\text{IoU}}{n}$, where $n$ is the total number of faraway objects and $\sum_{i=1}^n \text{IoU}$ is the sum of the IoU values calculated based on the ground truth and the predicted bounding box.

{\bf Implementation.} Our method uses the instance-level semantic segmentation method (Mask R-CNN~\cite{maskrcnn} pre-trained on COCO dataset~\cite{coco}) to generate 2D object information from image space.  
Our first approach (ours1) uses 2D semantic masks to generate frustums in 3D. The second approach (ours2) uses 2D bounding boxes to generate frustums. As a baseline (ours3), we also use ground truth 2D bounding boxes provided by KITTI to generate frustums. All of our approaches are combined with PV-RCNN~\cite{pvrcnn} for non-faraway objects. The Faraway Frustum Network (FF-Net) is trained using the Adam optimizer with early stopping. FF-Net is trained with the whole training set, but during inference only faraway objects are fed to the FF-Net.

% we use pre-defined values $w_i=0.7m, l_i=0.7m$, and $h_i=1.75m$ for pedestrians and $w_i=3m, l_i=3m$, and $h_i=1.5m$ for cars.

We compared our proposed method with the following SOTA 3D/BEV object detectors: SECOND~\cite{second}, PointPillars~\cite{pointpillars}, PV-RCNN~\cite{pvrcnn}, and Frustum PointNets~\cite{fp}. These SOTA methods are all trained using our data split from scratch and, we evaluated them with the same faraway metrics.

\section{RESULTS}

{\bf Quantitative results.} Table~\ref{table:iou} shows the average IoU results for BEV detection of faraway objects in the KITTI validation dataset. All of our methods outperform SOTA methods with higher average IoU (at least 0.051 and at most 0.157). And surprisingly, none of the methods except for ours can achieve an average of 0.1 IoU for both pedestrians and cars. This result not only demonstrates the effectiveness of our method, but also underlines an important shortcoming of SOTA methods. Furthermore, we believe finding the exact shape  of  faraway  objects is not a priority.  As  long  as  we  obtain  the 3D/BEV detection result with even a small IoU (e.g. 0.1), it can still be very useful for certain applications such as automated driving. In other words, a 0.1 IoU detection is better than a false negative. As such, we set IoU threshold to 0.1 for faraway objects in the mAP comparison.

%%%%%%%%%%% one colum
\begin{table}[htb]
\centering
\centering
\caption{Average IoU Comparison of Faraway BEV Object Detection on KITTI Val Dataset}
\begin{threeparttable}
\small
\renewcommand{\arraystretch}{1.1}
\begin{tabular}{p{3.5cm}<{\centering}||p{1.4cm}<{\centering}|p{1.4cm}<{\centering}}
\hline

\hline
% \multirow{Method} 

% & \multicolumn{1}{c|}{\begin{tabular}[c]{@{}c@{}} 3D Ped. \end{tabular}} 
% & \multicolumn{1}{c|}{\begin{tabular}[c]{@{}c@{}} BEV Ped. \end{tabular}} 
% & \multicolumn{1}{c|}{\begin{tabular}[c]{@{}c@{}} 3D Car \end{tabular}} 
% & \multicolumn{1}{c}{\begin{tabular}[c]{@{}c@{}} BEV Car \end{tabular}}  \\ \cline{2-5} 

\multirow{1}{*}{Method}
% \multirow{Method}
&\multicolumn{1}{c|}{\begin{tabular}[c]{@{}c@{}} BEV Ped.\end{tabular}}
&\multicolumn{1}{c}{\begin{tabular}[c]{@{}c@{}} BEV Car\end{tabular}} \\
\cline{2-3}

& \multicolumn{1}{c|}{\begin{tabular}[c]{@{}c@{}} $>$ 60 m \end{tabular}}
& \multicolumn{1}{c}{\begin{tabular}[c]{@{}c@{}} $>$ 75 m \end{tabular}}
\\
\cline{2-3}

\hline

\hline

Frustum PointNets \cite{fp}      &0.000    &0.000        \\ 

SECOND \cite{second}      &0.036 &0.009
  \\ 

PointPillars \cite{pointpillars}    &0.072   &0.000
  \\

PV-RCNN {\cite{pvrcnn}}    &0.051 &0.018 
 \\
\hline

Ours1 (mask + PV-RCNN)        &\textcolor[rgb]{0,0,1}{0.123}  &\bf {0.157}       \\

Ours2 (box + PV-RCNN)      & \bf 0.124       & \textcolor[rgb]{0,0,1}{0.150}  
  \\

%  \hline
 
% Ours3 (GT box + PV-RCNN)    & 73.27       &{69.67}   & {67.81} 
% & {\bf72.10}
% \\  
\hline

\hline

\end{tabular}

\begin{tablenotes}
        \footnotesize
        \item[*] Name explanation: Ped. (Pedestrian).
        \item[**] The {\bf bold} result means the best in all methods, and the {\textcolor[rgb]{0,0,1}{blue}} result represents the second place.
      \end{tablenotes}
    \end{threeparttable}

\label{table:iou}
\end{table}

%%%%%%%%%%%%%% 1st split  0902 pedestrian
%%%%%%%%%%%%%%%%%%%%%%%%%%%%%%%%%%%%%%%%%%%%%%%%%%%%%%%%%%%%%%%%%%%%%%%% 60m 0.5-0.1
\begin{table*}[ht]
\centering
\centering
\caption{ mAP Comparison of Faraway 3D/BEV Object Detection on KITTI Val Dataset}
\begin{threeparttable}
\small
\renewcommand{\arraystretch}{1.1}
\begin{tabular}{p{5cm}<{\centering}||p{1.4cm}<{\centering}|p{1.4cm}<{\centering}|p{1.4cm}<{\centering}|p{1.4cm}<{\centering} }
\hline

\hline
% \multirow{Method} 

& \multicolumn{1}{c|}{\begin{tabular}[c]{@{}c@{}} 3D Ped. \end{tabular}} 
& \multicolumn{1}{c|}{\begin{tabular}[c]{@{}c@{}} BEV Ped. \end{tabular}} 
& \multicolumn{1}{c|}{\begin{tabular}[c]{@{}c@{}} 3D Car \end{tabular}} 
& \multicolumn{1}{c}{\begin{tabular}[c]{@{}c@{}} BEV Car \end{tabular}}  \\ \cline{2-5} 

\multirow{1}{*}{Method}
&\multicolumn{4}{c}{\begin{tabular}[c]{@{}c@{}} IoU threshold 0.1 \end{tabular}}\\
\cline{2-5}

&\multicolumn{2}{c|}{\begin{tabular}[c]{@{}c@{}} Over 60 meters \end{tabular}}
& \multicolumn{2}{c}{\begin{tabular}[c]{@{}c@{}} Over 75 meters \end{tabular}} \\
\cline{2-5}

% & \multicolumn{1}{c||}{\begin{tabular}[c]{@{}c@{}} IoU thr. 0.1  \end{tabular}} 
% & \multicolumn{3}{c|}{\begin{tabular}[c]{@{}c@{}} IoU threshold 0.5 \end{tabular}} 
% & \multicolumn{1}{c}{\begin{tabular}[c]{@{}c@{}} IoU thr. 0.1  \end{tabular}} \\ \cline{2-9} 

% \multirow{Method} & \multicolumn{3}{c|}{\begin{tabular}[c]{@{}c@{}}AP on faraway pedestrian sub-dataset\\ {[}$\infty$,50m{]} \\ IoU threshold 0.25 / 0.01 \end{tabular}} & \multicolumn{3}{c}{\begin{tabular}[c]{@{}c@{}}AP on faraway pedestrian sub-dataset\\ {[}$\infty$,60m{]} \\ IoU threshold 0.25 / 0.01  \end{tabular}} \\ \cline{2-6} 
% \multirow{1}
% & \multicolumn{4}{c|}{\begin{tabular}[c]{@{}c@{}} IoU threshold 0.1 \end{tabular}}
% \cline{2-5}
% & Easy & Mod  & Hard   &     Over 60 m & Easy & Mod  & Hard  &     Over 60 m \\ 
\hline

\hline

Frustum PointNets \cite{fp}      &00.00    &00.00  &00.00 
&  00.00      \\ 

SECOND \cite{second}      &13.63 &13.63 &09.09 &09.09  
  \\ 

PointPillars \cite{pointpillars}    &22.40 &22.40 &00.00 &00.00
  \\

PV-RCNN {\cite{pvrcnn}}    &19.69 &19.69 &18.18 &18.18
 \\
\hline

Ours1 (mask + PV-RCNN)         &\bf {44.54}    &\textcolor[rgb]{0,0,1}{44.54} & \textcolor[rgb]{0,0,1}{34.70}  & \textcolor[rgb]{0,0,1}{45.27}   \\

Ours2 (box + PV-RCNN)          & \textcolor[rgb]{0,0,1}{31.95}  & \bf 45.45         & \bf {46.90} &\bf 46.90
  \\

%  \hline
 
% Ours3 (GT box + PV-RCNN)    & 73.27       &{69.67}   & {67.81} 
% & {\bf72.10}
% \\  
\hline

\hline

\end{tabular}

\begin{tablenotes}
        \footnotesize
        \item[*] Name explanation: Ped. (Pedestrian).
        \item[**] The {\bf bold} result means the best in all methods, and the {\textcolor[rgb]{0,0,1}{blue}} result represents the second place. We set the experimental IoU threshold as 0.1 for faraway pedestrians because in the current stage, it is extremely difficult to precisely locate faraway objects, while detecting faraway objects even with low IoU is still practical.
      \end{tablenotes}
    \end{threeparttable}

\label{table:ped1}
\end{table*}

%%%%%%%%%%%%%% 1st split  0902 easy.mod.hard
%%%%%%%%%%%%%%%%%%%%%%%%%%%%%%%%%%%%%%%%%%%%%%%%%%%%%%%%%%%%%%%%%%%%%%%% 75m 0.7-0.1
% table 3D/BEV
\begin{table*}[ht]
\centering
\centering
\caption{ mAP Comparison of 3D/BEV Object Detection on KITTI Val Dataset}
\begin{threeparttable}
\small

\renewcommand{\arraystretch}{1.1}
\begin{tabular}{p{3.5cm}<{\centering}||p{1.6cm}<{\centering}p{1.6cm}<{\centering}p{1.6cm}<{\centering}|p{1.6cm}<{\centering}p{1.6cm}<{\centering}p{1.6cm}<{\centering}}
\hline

\hline

& \multicolumn{3}{c|}{\begin{tabular}[c]{@{}c@{}} 3D/BEV Pedestrian \end{tabular}} 
& \multicolumn{3}{c}{\begin{tabular}[c]{@{}c@{}} 3D/BEV Car \end{tabular}} \\ \cline{2-7} 
% \multirow {Method} 
\multirow{1}{*}{Method}
& \multicolumn{3}{c|}{\begin{tabular}[c]{@{}c@{}} IoU threshold 0.5 \end{tabular}} 
% & \multicolumn{1}{c||}{\begin{tabular}[c]{@{}c@{}} IoU thr. 0.1  \end{tabular}}  
& \multicolumn{3}{c}{\begin{tabular}[c]{@{}c@{}} IoU threshold 0.7 \end{tabular}} 
% & \multicolumn{1}{c}{\begin{tabular}[c]{@{}c@{}}  IoU thr. 0.1  \end{tabular}} 
\\ \cline{2-7} 

% \multirow{Method} & \multicolumn{3}{c|}{\begin{tabular}[c]{@{}c@{}}AP on faraway pedestrian sub-dataset\\ {[}$\infty$,50m{]} \\ IoU threshold 0.25 / 0.01 \end{tabular}} & \multicolumn{3}{c}{\begin{tabular}[c]{@{}c@{}}AP on faraway pedestrian sub-dataset\\ {[}$\infty$,60m{]} \\ IoU threshold 0.25 / 0.01  \end{tabular}} \\ \cline{2-6} 

& Easy & Mod  & Hard   & Easy & Mod  & Hard   \\ 
\hline

\hline

PV-RCNN \cite{pvrcnn}  &69.53/{73.32} &66.02/67.42 &{{62.91}}/65.70 &{{96.73}}/97.53 &{93.18}/{94.77} &{85.76}/{94.78}
  \\

\hline

Ours1 (mask + PV-RCNN)    &71.65/73.02 &{67.29}/{68.73} &62.08/66.87 &{96.73}/{97.54} &{93.17}/{94.75} &{{85.76}}/{94.77} 
          \\ 
Ours2 (box + PV-RCNN)     &{{71.74}}/{73.11} &{67.29}/{68.73} &62.08/{66.88} &{96.73}/{97.54} &{93.17}/{94.75} &{{85.76}}/{94.77} 
      \\

\hline

\hline

\end{tabular}

\begin{tablenotes}
        \footnotesize
        \item[]  For the non-faraway official Easy/Mod/Hard benchmark, our method performs as well as the baseline SOTA method (PV-RCNN). This result shows that the proposed method can be used to improve the faraway detection performance without sacrificing the non-faraway detection performance.
      \end{tablenotes}
    \end{threeparttable}

\label{table:car}
\end{table*}

The mAP results of faraway 3D/BEV detection over KITTI validation dataset for pedestrians and cars are shown in Table~\ref{table:ped1}. For faraway pedestrians (over 60 meters), our methods (ours1 and ours2) outperform SOTA methods on 3D/BEV detection with large mAP margins (BEV: at least 22.14\% and at most 45.45\%, 3D: at least 9.55\% and at most 44.54\%). For faraway cars (over 75 meters), our methods (ours1 and ours2) outperform SOTA methods again with a higher mAP (BEV: at least 27.09\% and at most 46.90\%, 3D: at least 16.52\% and at most 46.90\%).

The mAP results of 3D/BEV detection over KITTI validation dataset for pedestrians and cars are shown in Table~\ref{table:car}. For the non-faraway official Easy/Mod/Hard benchmark, our method performs as well as the baseline SOTA method (PV-RCNN).

All the above results demonstrate that our method achieves better performance on faraway object detection without impairing the overall performance of SOTA methods. 

{\bf Qualitative results.} Fig. \ref{fig:example} shows an example of visual results of different methods for faraway object detection. We compared our method with PV-RCNN~\cite{pvrcnn} and Frustum PointNets~\cite{fp}. In frame (a), Frustum PointNets mistakenly detects the pole as a pedestrian, and PV-RCNN even has no result. However, for detecting the faraway pedestrian near the pole, only our detector succeeds. In frame (b), state-of-the-art methods all fail in detecting the faraway car. In contrast, our method successfully detects that car in 3D.

%%%%%%%%% Conclusion
\section{CONCLUSION}

In this paper, we proposed an alternative 3D/BEV detector, named {\textit{Faraway-Frustum}}, to deal with lidar sparsity of faraway objects. Our method takes the advantages of relatively dense image data to find faraway objects, and circumvents the disadvantages of pointcloud-driven neural networks working on very sparse points. Moreover, our alternative detector can be flexibly combined with a state-of-the-art method to form an overall 3D/BEV object detection system via setting faraway thresholds. 

The experiments demonstrated the feasibility of our approach, but they also exposed a significant shortcoming of state-of-the-art object detection methods: Relying on learned representations of very sparse lidar points to detect faraway objects is not a good strategy.

\section*{ACKNOWLEDGMENT}

Material reported here was supported by the United States Department of Transportation under Award Number 69A3551747111 for the Mobility21 University Transportation Center.

%%%%%%%%%%%%%%%%%%%%%%%%%%%%%%%%%%%%%%%%%%%%%%%%%%%%%%%%%%%%%%%%%%%%%%%%%%%%%%%%

\bibliographystyle{IEEEtran}
\bibliography{IEEEabrv,ref}

\end{document}